\documentclass{article}
\usepackage{spconf,amsmath,graphicx,cases}
\usepackage{comment}
\usepackage{url}
\newcommand{\etal}{\textit{et al}.}

\title{Toward Unsupervised 3D Point Cloud Anomaly Detection \\ using Variational Autoencoder}

\name{Mana Masuda$^{\star}$\qquad Ryo Hachiuma$^{\star}$\qquad Ryo Fujii$^{\star}$\qquad Hideo Saito$^{\star}$\qquad Yusuke Sekikawa$^{\dagger}$}
\address{$^{\star}$ Keio University, Japan \\ 
    $^{\dagger}$Denso IT Laboratory, Japan}

\begin{document}

\maketitle

\begin{abstract}
    In this paper, we present an end-to-end unsupervised anomaly detection framework for 3D point clouds. To the best of our knowledge, this is the first work to tackle the anomaly detection task on a general object represented by a 3D point cloud. We propose a deep variational autoencoder-based unsupervised anomaly detection network adapted to the 3D point cloud and an anomaly score specifically for 3D point clouds. To verify the effectiveness of the model, we conducted extensive experiments on the ShapeNet dataset. Through quantitative and qualitative evaluation, we demonstrate that the proposed method outperforms the baseline method.
    Our code is available at \url{https://github.com/llien30/point_cloud_anomaly_detection}.
\end{abstract}

\vspace{-2mm}
\section{Introduction}
\label{sec:intro}
Anomaly detection is the task of recognizing whether an input sample is within the distribution of a given target normal class or an anomaly class. Anomaly detection is a fundamental task in various fields, such as detecting malicious actions, system failures, intentional fraud, and diseases. Many deep learning-based methods have been proposed \cite{chalapathy2019deep}, with a wide range of input data, including sound \cite{suefusa2020anomalous}, big data \cite{8373692}, signal data \cite{ahrens2019machine}, natural language \cite{DBLP:journals/corr/abs-1908-09156}, image \cite{schlegl2017unsupervised}, and video \cite{park2020learning}.

Thanks to the development of 3D sensing devices, such as LiDAR, stereo cameras, and structured light sensors, 3D point clouds are ubiquitous today. As a result, there has been growing interest in developing algorithms for performing classification \cite{Yan_2020_CVPR}, segmentation \cite{Yan_2020_CVPR}, and object detection \cite{Shi_2020_CVPR}. Unlike images, 3D data can be represented in various ways, such as 3D volumes, meshes, and point clouds (set of points). Based on PointNet~\cite{qi2017pointnet}, many methods for processing point clouds have been proposed that handle the permutation invariance of the input data and memory efficiency. 
Sekuboyina \etal~\cite{sekuboyina2019probabilistic} proposed an anomaly detection method for 3D point clouds for analyzing vertebral shapes, but this method is not suitable for detecting the anomaly of general objects, as the network can reconstruct only a fixed number of point data.

In this paper, we tackle the task of detecting anomalies for the 3D point clouds of a general object in an unsupervised manner. A formal definition of an unsupervised anomaly detection task is as follows: given a dataset $\mathcal{D}$ that contains a large number of normal data $X$ for training, and several abnormal data $\hat{X}$ for testing, model $f$ is optimized over its parameter $\theta$ using training data $X$. $f$ learns normal distribution $p_x$ during training and identifies abnormal data as outliers during testing by outputting an anomaly score $\mathcal{A}(x)$, where $x$ is a given test sample. A larger $\mathcal{A}(x)$ indicates possible abnormalities within the test sample because $f$ learns to minimize the output score during training.

In this paper, we present a variational autoencoder (VAE)-based anomaly detection method for  3D point clouds. Considering the characteristics of the anomaly detection task, we hypothesized that the anomaly detection task in 3D point clouds should also be solved with a reconstruction-based method, referring to the anomaly detection method for images \cite{schlegl2017unsupervised, akcay2018ganomaly, kimura2020adversarial, venkatakrishnan2020self}. Although many reconstruction methods have been proposed for 3D point clouds \cite{qi2017pointnet, yang2018foldingnet, deng2018ppf}, they are not adapted to the anomaly detection task.

In addition, we have conducted many experiments to evaluate the loss during training and the method for measuring the anomaly score $A(x)$ of 3D point clouds, which is appropriate for solving the task of anomaly detection in 3D point clouds. To validate the proposed method, we performed a category-out experiment on the ShapeNet dataset \cite{shapenet2015}, referring to the experiments of anomaly detection methods on images \cite{akcay2018ganomaly, kimura2020adversarial}.

\begin{figure*}[tb]
\centering
\includegraphics[width=0.8\hsize]{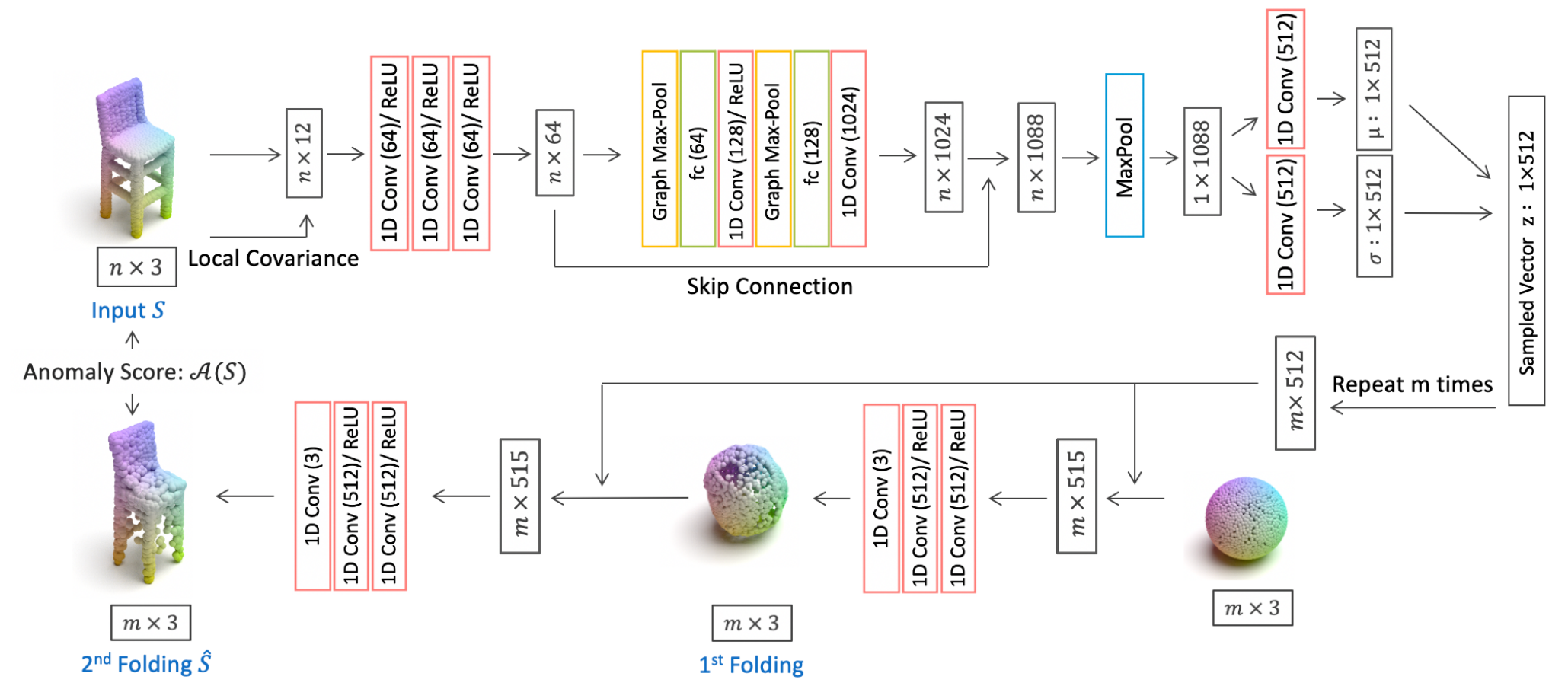}
\caption{Overview of the proposed method. We adopted a FoldingNet-based decoder \cite{yang2018foldingnet} and introduced a skip connection in the encoder, which allows compressed features to include global and local features.}
\label{fig:network overview}
\end{figure*}

The contributions of the paper are as follows:
\begin{itemize}
    \item As far as we know, this paper is the first to tackle the anomaly detection task for 3D point clouds of general objects. We present an anomaly detection framework based on a variational autoencoder. We also present the loss function and the anomaly score calculation for 3D point clouds. 
    \item We conducted extensive experiments to validate the proposed network, loss function, and anomaly score. The results verified that the proposed method achieves high accuracy of more than $76$\% on average when the area under the curve (AUC) of the receiver operating characteristic (ROC) is used as an evaluation metric. 
\end{itemize}

\section{Methods}
\label{sec:methods}
An overview of the proposed method is shown in Fig. \ref{fig:network overview}. Inspired by the conventional anomaly detection method for images \cite{schlegl2017unsupervised, akcay2018ganomaly, kimura2020adversarial, venkatakrishnan2020self}, we propose a reconstruction-based anomaly detection method for 3D point clouds. At training time, given normal point clouds as inputs, the model extract feature distributions and try to reconstruct the inputs. The point clouds of large reconstruction errors are then treated as anomalies at test time. This method is also inspired by the feature learning method (FoldingNet) \cite{yang2018foldingnet}. 

\subsection{Model Overview}
\label{sec:model_overview}
We propose a VAE model suitable for anomaly detection of 3D point clouds. For the encoder, we introduce skip-connection and a graph max-pooling layer which estimates local features based on the graph structure \cite{Shen_2018_CVPR}. For the decoder, we use the FoldingNet \cite{yang2018foldingnet} decoder, but we adopt a spherical shape as the grid instead of a plane. The input for the encoder is an $n$-by-$3$ matrix. Each row of the matrix is composed of the 3D position $(x,y,z)$. The encoder concatenates the local covariance matrix proposed by Yang \etal~\cite{yang2018foldingnet} to the input before input to the convolution layer. The output is also an $n$-by-$3$ matrix representing the reconstructed point positions. The encoder computes the mean $\mu$ and variance $\sigma$ from each input point cloud, and the decoder reconstructs the point cloud using the sampled vector $z$ from its mean $\mu$ and variance $\sigma$. The mean $\mu$ and variance $\sigma$ length is set as $512$ in accordance with Achlioptas \etal~\cite{achlioptas2018learning}.

\subsection{Model Training}
\label{sec:training}
\subsubsection{Reconstruction loss}
Two permutation-invariant metrics for comparing unordered point sets have been proposed \cite{fan2017point}. For the reconstruction error between the original point cloud $S$ and the reconstructed point cloud $\hat{S}$, the earth mover's distance (EMD)\cite{rubner2000earth},
\begin{equation}
\label{eq:emd}
    d_{EMD}(S, \hat{S}) = \min_{\phi:S \to \hat{S}} \sum_{x \in S} ||x-\phi (x)||_2,
\end{equation}
where $\phi:S \to \hat{S}$ is a bijection and the Chamfer distance (CD),
\begin{equation}
\label{eq:cd}
    \begin{split}
    &d_{CD}(S, \hat{S}) = \\
    & \quad \frac{1}{|S|}\sum_{x\in S}\min_{\hat{x}\in\hat{S}}||x-\hat{x}||_2 + \frac{1}{|\hat{S}|}\sum_{x\in \hat{S}}\min_{x\in S}||\hat{x}-x||_2,
    \end{split}
\end{equation}
can be considered. Following FoldingNet \cite{yang2018foldingnet}, we employ the CD for the reconstruction error $\mathcal{L}rec$, because training with the CD is faster in terms of convergence, and the CD is less computationally expensive than the EMD. 

\subsubsection{KL divergence}
Following the traditional method of the VAE \cite{kingma2013auto}, we adopt the KL divergence as a loss. We compute the loss to be small for the KL divergence between Gaussian distribution $\mathcal{N}(0,1)$ and $\mathcal{N}(\mu, \sigma)$ computed from the original point cloud $S$. We define this KL divergence
as $D_{KLori}$:
\begin{equation}
    D_{KLori} = D_{KL}(\mathcal{N}(\mu,\sigma^2)||\mathcal{N}(0,1)).
\end{equation}

We also adopt the second KL divergence between Gaussian distribution $\mathcal{N}(0,1)$ and $\mathcal{N}(\hat{\mu},\hat{\sigma})$, where $\hat{\mu}$ and $\hat{\sigma}$ are obtained by entering the reconstructed point cloud $\hat{S}$ into the network. We define this KL divergence 
as $D_{KLrec}$:

\begin{equation}
    D_{KLrec} = D_{KL}(\mathcal{N}(\hat{\mu}, \hat{\sigma}^2)||\mathcal{N}(0,1)).
\end{equation}

\begin{figure*}[tb]
\centering
\includegraphics[width= 0.85\hsize]{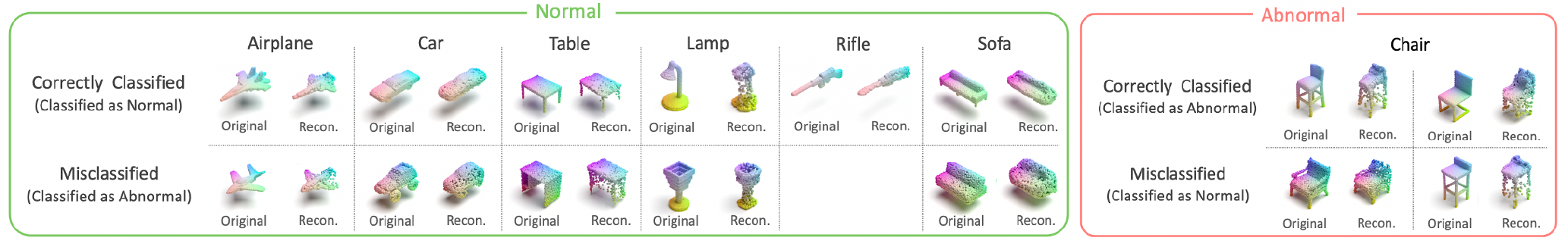}
\caption{Qualitative results when the chair is the anomaly class. The top row shows the correctly classified samples, and the bottom row shows the incorrectly classified samples. The left side of each sample is the original point cloud and the right side is the reconstructed point cloud. The rifles had small reconstruction errors, and the proposed model did not misclassify them as abnormal.}
\label{fig:experiment}
\end{figure*}

Overall, the objective function becomes the following:
\begin{equation}
    \mathcal{L}  = \mathcal{L}rec + D_{KLori} + D_{KLrec}.
\end{equation}

\subsection{Anomaly Detection}
\label{sec:anomaly_detection}
To measure whether the sample is anomalous or not, we adapt the anomaly score proposed in \cite{schlegl2017unsupervised}. We choose the Chamfer distance as the anomaly score $\mathcal{A}(S)$ for 3D point clouds.

\section{Experiment and Results}
\label{sec:experiments}
To evaluate the anomaly detection framework, we used the ShapeNet dataset~\cite{shapenet2015}. We used only seven classes of the datasets (airplane, car, chair, lamp, rifle, table, sofa) that included more than $2000$ data. All data were pre-processed by sampling $2048$ points randomly. During training, we set the number of points in the output layer to $2048$ ($m=2048$). To validate the proposed method, we performed a category-out experiment referring to image-based anomaly detection methods \cite{akcay2018ganomaly, kimura2020adversarial}. To verify the anomaly detection method for 3D point clouds, we conducted two different experiments: The first was a comparison of the models, and the second was a comparison of the anomaly scores. Mitsuba2 renderer \cite{nimier2019mitsuba} was used to visualize the dataset and its reconstruction results. We implemented the code by regarding this repository\footnote{\url{https://github.com/AnTao97/UnsupervisedPointCloudReconstruction}}.

\subsection{Quantitative Evaluation}
\label{sec: model_comparison}
We compared the proposed model with an anomaly detection result with FoldingNet and summarized the results in Table \ref{table: model_comparison}. Following the experimental setup in \cite{akcay2018ganomaly, kimura2020adversarial}, we measured the average AUC by computing the area under the ROC with varying threshold values for the anomaly scores. We report the AUC performance of two of the models, with and without $D_{KLrec}$ in the loss. From the table, we verify two things: (1) The model without $D_{KLrec}$ in the loss shows a better result than the FoldingNet \cite{yang2018foldingnet}, achieving an average AUC of $75.1$\%. This demonstrates the effectiveness of the proposed VAE network for anomaly detection. (2) The model with $D_{KLrec}$ in the loss shows the best result among the three models, achieving an average AUC of 76.3\%. Although we did not use this value as the anomaly score, the anomaly detection accuracy was improved when $D_{KLrec}$ was used as the loss. This indicates that $D_{KLrec}$ is a very effective loss for the reconstruction of 3D point clouds for anomaly detection.

\begin{table}[tb]
\begin{center}
\caption{Quantitative comparison of an anomaly detection result with FoldingNet \cite{yang2018foldingnet}. We measured the AUC (\%) on the ShapeNet dataset \cite{shapenet2015}. 
Numbers in bold indicate the best performance, and underscored numbers are the second best. We set the number of output points to $4096$.}
\label{table: model_comparison}
\small
\begin{tabular}{l|c|c|c}
\hline
         & FoldingNet \cite{yang2018foldingnet} & w/o $D_{KLrec}$ &    w/ $D_{KLrec}$(ours)\\ \hline
airplane &                 72.9                 &   \bf{77.0}     &   \underline{74.7}   \\
car      &              \bf{83.0}               &      72.4       &   \underline{75.7}   \\
chair    &                 88.9                 &\underline{89.5} &       \bf{93.1}      \\
lamp     &                 84.2                 &\underline{90.1} &       \bf{90.7}      \\
table    &                 77.7                 &   \bf{87.1}     &   \underline{83.9}   \\
rifle    &                 25.7                 &\underline{33.0} &       \bf{38.2}      \\
sofa     &                 73.6                 &\underline{76.5} &       \bf{77.7}      \\ \hline
average &                 72.3                 &\underline{75.1} &       \bf{76.3}      \\ \hline
\end{tabular}
\end{center}
\end{table}

\begin{table*}[tb]
\begin{center}
\caption{Anomaly score ablation study. When we added two anomalies together,  we normalized each anomaly score to a magnitude between zero and one. As in Table~\ref{table: model_comparison}, we measured the AUC (\%) on ShapeNet dataset \cite{shapenet2015} for each anomaly score. Numbers in bold indicate the best performance. The number of output points is fixed to $2048$ for comparison with the EMD.}
\label{table:anomaly_score_ablation}
\small
\begin{tabular}{lccccccc|c}
\hline
                     &    airplane   &     car    &    chair   &     lamp   &   table    &    rifle   &    sofa    &   average  \\ \hline
$||(\mu+\epsilon \times \sigma) - (\hat{\mu}+ \hat{\epsilon} \times \hat{\sigma})||_2$  &     50.5     &    44.7   &    30.3   &    53.4   &   45.7    &    37.2   &    47.0   &    44.1   \\ \hline
$D_{KL}(\mathcal{N}(\mu, \sigma) || \mathcal{N}(0,1) )$                                  &     36.9     &    70.2   &    94.7   &    57.6   &   89.0    &    34.4   &    52.1   &    62.1   \\ \hline
$N_{scale}(d_{EMD}(S, \hat{S}))+N_{scale}(D_{KL}(\mathcal{N}(\mu, \sigma)) || \mathcal{N}(0,1))$             &     57.5     &    59.3   &    91.5   &    76.1   &   88.8    &    37.3   &    57.2   &    66.8   \\ \hline
$d_{EMD}(S, \hat{S})$                                                                   &     65.4     &    49.3   &    78.4   &    86.4   &   82.1    & \bf{44.4} &    62.2   &    66.9   \\ \hline
$N_{scale}(d_{CD}(S, \hat{S}))+N_{scale}(D_{KL}(\mathcal{N}(\mu, \sigma)) || \mathcal{N}(0,1))$              &     56.9     &    73.7   & \bf{96.3} &    73.7   & \bf{90.3} &    29.9   &    69.4   &    70.0   \\ \hline
$d_{CD}(S, \hat{S})$                                                                    &  \bf{71.6}   & \bf{75.2} &    91.8   & \bf{90.3} &   83.4    &    28.6   & \bf{77.8} & \bf{74.1} \\ \hline

\end{tabular}
\end{center}
\end{table*}

\subsection{Qualitative Evaluation}
\label{sec: qualitatibe_eval}
In Fig. \ref{fig:experiment}, we show the qualitative results of the proposed model when the chair is the anomaly class. Fig. \ref{fig:experiment} shows an example in which normal and abnormal samples are correctly and incorrectly classified, respectively, based on the threshold value calculated so that the best accuracy would be achieved. The top row shows the correctly classified samples, and the bottom row shows the misclassified samples. The left side of each sample is the original point cloud and the right is the reconstructed point cloud. From the qualitative results, it can be confirmed that for normal data, data that are correctly reconstructed are correctly classified as normal, while data that are poorly reconstructed are incorrectly classified as abnormal. For abnormal data, data with relatively good reconstructions were misclassified as normal data, and data with poor reconstructions were correctly classified as abnormal data.

\begin{figure}[tb]
\centering
\includegraphics[width=0.8\hsize]{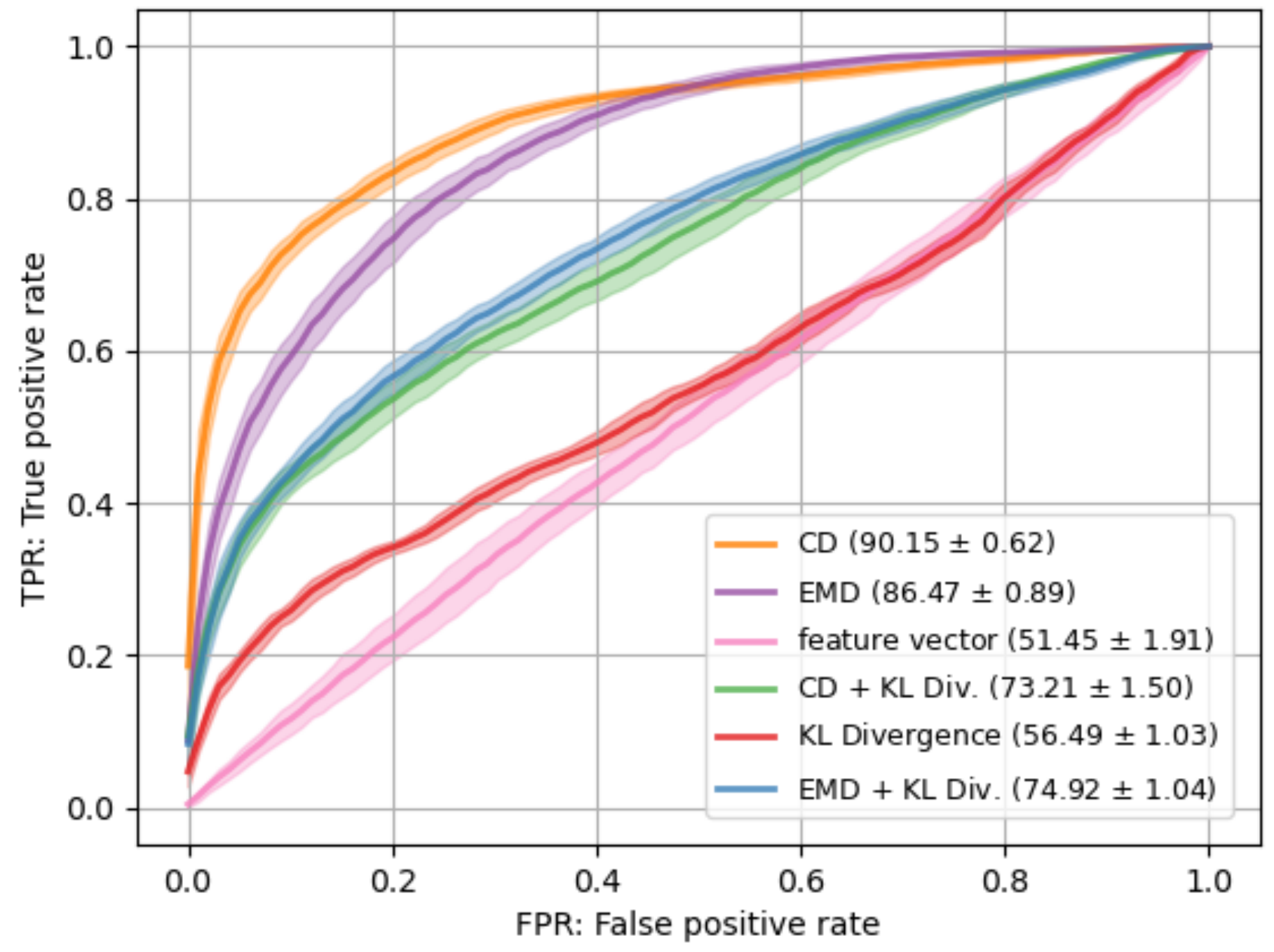}
\caption{Comparison of ROC curves for the average of $50$ random seeds of the six anomaly scores with lamps as the anomaly class. Shaded areas in the plot represent the variance.}
\label{fig:roc_curve}
\end{figure}

\subsection{Ablation Study of the Anomaly Score}
\label{sec: anomaly_score_comparison}
Because this is the first work that tackles the anomaly detection task on a general object represented by a point cloud, an ablation study of various scores was conducted. There are mainly two types of anomalies in the point cloud: reconstruction errors and feature differences. For the reconstruction error, we considered two sorts of errors: the EMD and the CD. For the feature difference, we considered two types of feature difference: the KL divergence between the unit Gaussian and the Gaussian predicted from the input point clouds and the sampled feature from Gaussian distribution. We considered that reconstruction errors and feature differences are important, and examined the five anomaly scores in addition to the proposed anomaly score. We report the AUC performance for the variants of the anomaly scores of the proposed model in Table \ref{table:anomaly_score_ablation}. As there is a numerical difference between the reconstruction error and the difference in the feature values, when we added the two types of anomaly scores, we scaled each anomaly score to one at the maximum and zero at the minimum:
\begin{equation}
    N_{scale}(X) = \frac{X - x_{min}}{x_{max}-x_{min}},
\end{equation}
where $x_{min}$ and $x_{max}$ are the minimum and maximum anomaly scores in $X$, and then added them for the comparison. Fig. \ref{fig:roc_curve} shows the ROC curves and their variance for the six anomaly scores with lamps as the anomaly class. When we used only the CD as the anomaly score, not only was the AUC score the best but also the variance was the smallest, indicating that the accuracy was stable and the best.

\begin{table}[tb]
\begin{center}
\caption{Comparison of the number of output points. We measured the accuracy with AUC (\%). Numbers in bold indicate the best performance.}
\label{table: num_point_comparison}
\small
\begin{tabular}{l|c|c|c|c|c}
\hline
         & 1024 &   2048  &   3072   &   4096   &   5120   \\ \hline
airplane & 58.7 &   71.6  &   72.0   &   74.7   & \bf{75.5}\\
car      & 70.5 &   75.2  &   75.4   & \bf{75.7}&   75.4   \\
chair    & 87.8 &   91.8  &   92.6   & \bf{93.1}& \bf{93.1}\\
lamp     & 87.2 &   90.3  &   90.6   &   90.7   & \bf{90.8}\\
table    & 80.2 &   83.4  &   83.2   & \bf{83.9}&   83.4   \\
rifle    & 17.8 &   28.6  &   31.7   & \bf{38.2}&   37.4   \\
sofa     & 75.1 &   77.8  &   77.8   &   77.7   & \bf{78.1}\\ \hline
average  & 68.2 &   74.1  &   74.8   & \bf{76.3}&   76.2   \\ \hline
\end{tabular}
\end{center}
\end{table}

\subsection{Ablation Study of the Number of Points}
Table \ref{table: num_point_comparison} shows a comparison of the accuracy according to the number of reconstructed points. The number of points with the highest accuracy depends on the shape, but on average, the highest accuracy was $76.3$\% at $4096$ points. From the table, we can confirm that the effective number of reconstructed points differed among the objects' categories. 

\section{Conclusions}
\label{sec:conclusion}
We presented a novel unsupervised method for 3D point cloud anomaly detection. We evaluated a deep variational autoencoder network and a loss function and showed that both are suitable for 3D point cloud anomaly detection. We also compared various anomaly score functions and their combinations for anomaly detection in 3D point clouds and proposed the optimal anomaly score for 3D point clouds. In the future, we will tackle the practical application of anomaly detection of 3D point clouds. For example, the 3D point cloud anomaly detection task is useful for finding defects in industrial products.

\bibliographystyle{IEEEbib}
\bibliography{refs}

\end{document}